\documentclass[12pt,a4paper]{article}

\usepackage[T1]{fontenc}
\usepackage{lmodern}

\usepackage[
  top=2.5cm, bottom=2.5cm, left=2.5cm, right=2.5cm
]{geometry}

\usepackage[english]{babel}
\usepackage{colortbl}
\usepackage{tipa}
\usepackage{float}
\usepackage{graphicx}
\usepackage{arydshln}
\usepackage{xcolor}
\definecolor{newhighlight}{HTML}{DFF0D8}

\usepackage{amsmath}
\usepackage{microtype}
\usepackage{setspace}
\usepackage{tikz}
\usepackage{pgfplots}
\pgfplotsset{compat=1.8}
\usepackage{titlesec}
\titleformat{\section}{\large\bfseries}{\thesection}{1em}{}
\titleformat{\subsection}{\normalsize\bfseries}{\thesubsection}{1em}{}
\titleformat{\subsubsection}{\normalsize\bfseries}{\thesubsubsection}{1em}{}

\usepackage[bottom]{footmisc}

\usepackage{booktabs}          % Professional tables
\usepackage{graphicx}
\usepackage{caption}
\usepackage{csquotes}
\usepackage[
  backend=biber,
  style=unified,
  maxcitenames=2,              % "et al." for 3+ authors in citations
  maxbibnames=99,              % List ALL authors in reference entries
]{biblatex}
\usepackage[hidelinks]{hyperref}
\usepackage{cleveref}
\crefname{section}{Section}{Sections}
\crefname{table}{Table}{Tables}
\crefname{figure}{Figure}{Figures}

\newcommand{\shorttitle}{}     % Max 40 characters for running head
\newcommand{\authororcid}[1]{\textsuperscript{\,\href{https://orcid.org/#1}{ORCID}}}

\usepackage{fancyhdr}
\begin{document}

\title{Transformers over-extend what humans underlearn: the case of Spanish L-shaped morphome}
\author{%
Akhilesh Kakolu Ramarao\textsuperscript{1} \and
Kevin Tang\textsuperscript{1,3,*} \and
Dinah Baer-Henney\textsuperscript{2} \\[1ex]
{\small\textsuperscript{1}Department of English Language and Linguistics, Heinrich Heine University D\"{u}sseldorf} \\
{\small\textsuperscript{2}Institut f\"{u}r Germanistik, Philologische Fakult\"{a}t, Ruhr-Universit\"{a}t Bochum} \\
{\small\textsuperscript{3}Department of Linguistics, College of Liberal Arts and Sciences, University of Florida} \\[0.5ex]
{\small\textsuperscript{*}Corresponding author: \texttt{kevin.tang@uni-duesseldorf.de}}%
}
\date{}
\maketitle
\thispagestyle{fancy}

% --- Abstract and keywords ---------------------------------------------------
\begin{abstract}
\noindent
The cognitive reality of irregular morphological patterns has been debated for decades: do speakers extend them to novel forms, or are they lexical artifacts? A neural network trained on distributional input offers a learnability test: if it recovers the pattern, the pattern is learnable from input statistics alone. We apply this test to the Spanish L-shaped morphome, where the first-person singular indicative stem appears in every present subjunctive cell despite lacking apparent phonological or semantic motivation. We further ask whether the frequency of irregular verbs in the input modulates generalization, evaluating transformers under three frequency conditions (10\%, 50\%, 90\% irregular) and comparing them to human behavioral data. On full-form production from pseudoword inputs all models performed poorly, but all three conditions produced the correct stem more often than humans (43--49\% vs. 33\%). Response preferences revealed a clear divergence: humans consistently favored regular inflections, whereas models preferred irregular forms more as their proportion in training grew. Models in the naturalistic and balanced conditions were also sensitive to phonological similarity between pseudowords and real Spanish irregular verbs, an effect absent in humans. The L-shaped morphome is thus learnable from distributional input alone, but models generalize it qualitatively differently from humans.
\end{abstract}


\section{Introduction}
\label{sec:intro}

The cognitive status of morphomic patterns has been an open question in morphological theory ever since \textcite{Aronoff-1994} introduced the \textit{morphome}: a systematic mapping between arbitrary classes of morphosyntactic features (e.g., tense, number) and arbitrary sets of morphophonological forms (e.g., vowel alternations, suppletive stems) lacking transparent semantic or phonological motivation -- a pattern documented across a wide range of languages \cite{Herce2023}. Whether speakers extend these patterns productively to novel forms or memorize them on an item-by-item basis remains unresolved, and the answer bears directly on how morphological knowledge is organized in the grammar \parencite{maiden2013latin, otero2016, herce2022quantifying, herce2024meaning}.

We focus on the Spanish \textit{L-shaped morphome}, first identified by \textcite{Maiden1992} where the same stem appears in 1\textsc{sg.ind} and all cells of the present subjunctive (Table \ref{tab:paradigms}). Regular Spanish verbs, by contrast, do not undergo such stem alternation \parencite{raeconjugation2025}. The L-shaped is a minority pattern in Spanish: roughly 7\% of verbs exhibit it \textcite{ramarao-etal-2025-frequency}, a distribution that motivates our frequency manipulation (Section \ref{sec:methods}). 

\begin{table}[ht]
\centering
\renewcommand{\arraystretch}{1.3}
\resizebox{0.7\textwidth}{!}{
\begin{tabular}{l ll @{\hspace{2em}} ll}
\toprule
& \multicolumn{2}{c}{\textit{salir} `to leave' (irregular)} & \multicolumn{2}{c}{\textit{comer} `to eat'
(regular)} \\
\cmidrule(lr){2-3} \cmidrule(lr){4-5}
& \textbf{Indicative} & \textbf{Subjunctive} & \textbf{Indicative} & \textbf{Subjunctive} \\
\midrule
\textbf{1\textsc{sg}} & \cellcolor{newhighlight}\textbf{salg}-\textipa{o}    &
\cellcolor{newhighlight}\textbf{salg}-\textipa{a}    & \textipa{kom}-\textipa{o}    &
\textipa{kom}-\textipa{a}    \\
\textbf{2\textsc{sg}} & \textipa{sal}-\textipa{es}                           &
\cellcolor{newhighlight}\textbf{salg}-\textipa{as}   & \textipa{kom}-\textipa{es}   &
\textipa{kom}-\textipa{as}   \\
\textbf{3\textsc{sg}} & \textipa{sal}-\textipa{e}                            &
\cellcolor{newhighlight}\textbf{salg}-\textipa{a}    & \textipa{kom}-\textipa{e}    &
\textipa{kom}-\textipa{a}    \\
\textbf{1\textsc{pl}} & \textipa{sal}-\textipa{imos}                         &
\cellcolor{newhighlight}\textbf{salg}-\textipa{amos} & \textipa{kom}-\textipa{emos} &
\textipa{kom}-\textipa{amos} \\
\textbf{2\textsc{pl}} & \textipa{sal}-\textipa{is}                           &
\cellcolor{newhighlight}\textbf{salg}-\textipa{ajs}  & \textipa{kom}-\textipa{ejs}  &
\textipa{kom}-\textipa{ajs}  \\
\textbf{3\textsc{pl}} & \textipa{sal}-\textipa{en}                           &
\cellcolor{newhighlight}\textbf{salg}-\textipa{an}   & \textipa{kom}-\textipa{en}   &
\textipa{kom}-\textipa{an}   \\
\bottomrule
\end{tabular}
}
\caption{Present tense paradigms of an irregular L-shaped verb (\textit{salir} `to leave') and a regular verb (\textit{comer} `to eat'), given in IPA (Argentinian Spanish). Hyphens mark stem–suffix boundaries. For \textit{salir}, bold marks the L-shaped stem \textbf{salg-}, which recurs in 1\textsc{sg.ind} and every
subjunctive cell (shaded); all other cells use the same stem \textit{sal-}. For \textit{comer}, the stem \textit{kom-} is same throughout.}
\label{tab:paradigms}                  
\end{table}

The empirical evidence is contested. In a pseudoword production study, \textcite{Nevins2015TheRA} showed Spanish speakers inflected forms of each pseudoword -- one filling an L-shaped paradigm cell (e.g., 1\textsc{sg.ind}) and one filling a non-L-shaped cell (e.g., 2\textsc{sg.ind}) -- and asked them to produce a third cell (see Figure \ref{fig:stimuli-nevins-et-al}). Because the L-shaped stem was always visible in one of the two given forms, a speaker who has internalized the L-shaped pattern should select it in the target cell in nearly all responses. Instead, participants largely failed to generalize: only 33\% of responses used the L-shaped stem. An analogous Italian study \parencite{cappellaro2024cognitive}, using a forced-choice task in which participants selected between L-shaped and non-L-shaped forms, reported the opposite pattern: roughly 60\% of Italian speakers preferred the L-shaped alternative, suggesting that the pattern might be cognitively real in Italian.

This debate motivates a complementary approach. Computational models offer a way to test learnability hypotheses that are difficult to manipulate in human experiments \parencite{Dupoux2018, keller-2010-cognitively}. We reframe the cognitive reality of morphomes as a problem of \textit{computational learnability}: if a model trained solely on distributional input generalizes the L-shaped pattern, this constitutes evidence that the pattern is learnable from the statistical properties of the language; if it does not, this suggests that additional cognitive or experiential factors are needed. Recent work shows that neural networks trained on ecologically plausible input develop linguistic competences that track human performance on morphological tasks \parencite{warstadt-etal-2023-findings, evanson-etal-2023-language, Wilcox2024, kallini-etal-2024-mission}. 

A critical factor in morphological productivity for humans is type frequency \parencite{bybee1995,pierrehumbert2001stochastic,bybee2003phonology,Albright2003,baer2012role,guzmannaranjo2021coding}. To enable direct comparison between model and human performance, we train transformer models on Spanish verbs and evaluate them on the pseudoword stimuli used in \textcite{Nevins2015TheRA} study. Following \textcite{ramarao-etal-2025-frequency}, we train transformer models on three frequency conditions that vary the proportion of L-shaped verbs in the training data: (i) a naturalistic \textit{10\%L-90\%NL} condition that approximates the real Spanish distribution; (ii) a balanced \textit{50\%L-50\%NL} split; and a (iii) reversed \textit{90\%L-10\%NL} condition. We evaluate these models using the same test items and analytical framework as the \textcite{Nevins2015TheRA} human study, allowing direct model-human comparison. While statistical alignment between models and behavioral responses is a necessary baseline for cognitive plausibility, it does not by itself confirm human-like mechanisms \parencite{Guest2023}. The frequency manipulation allows us to go further and test whether the models track the distributional properties that drive human sensitivity.

We address three research questions:

\begin{itemize}
    \item \textbf{RQ1} Is the L-shaped morphome learnable from distributional input alone, and does the proportion of L-shaped verbs in training modulate generalization?
    \item \textbf{RQ2} How do model response preferences compare to human patterns reported by \textcite{Nevins2015TheRA}?
    \item \textbf{RQ3} Are models and humans more likely to extend the L-shaped pattern to pseudoword items that are phonologically similar to real Spanish verbs?
\end{itemize}

\section{Background}
\label{sec:background}

A central question in the study of morphological learning is how speakers learn and extend both regular and irregular inflectional patterns from the same input. This question has been debated for decades, with the English past tense as the primary testing ground. \textcite{rumelhart1986}'s pioneering connectionist network learned regular and irregular past-tense forms with modest success (\~67\% on stems), but \textcite{pinker1988language} argued its systematic errors showed a single mechanism could not cover both regular and irregular inflection. 

Modern neural networks have revived interest in this task. Sequence-to-sequence models have been applied to morphological inflection across typologically diverse languages such as Finnish, Arabic and Russian \parencite[\textit{inter alia}]{cotterell-etal-2017-conll,cotterell-etal-2018-conll,mccarthy-etal-2019-sigmorphon,vylomova-etal-2020-sigmorphon,pimentel-ryskina-etal-2021-sigmorphon,kodner-khalifa-2022-sigmorphon,goldman2023sigmorphon}. Whether such models generalize the same way as human speakers is sensitive to architectural choices and training regimes  \parencite{payne2025innate, kodner2025evaluating, heitmeier2025deeper, prickett2025learning, jarosz2025incremental, divjak2026synergy, ramarao-etal-2026-character}, making direct comparison with behavioral data essential.

So far, such comparisons have yielded mixed results. \textcite{kirov-cotterell-2018-recurrent} showed that a modern encoder-decoder addresses most of \textcite{pinker1988language}'s original criticisms, but \textcite{corkery-etal-2019-yet} found that such models correlate only weakly with human judgments on pseudoword verbs ($\rho = 0.15$--$0.56$ for regulars, $0.23$--$0.41$ for irregulars) and fail to capture the categorical regular-irregular distinction. Similarly, \textcite{mccurdy2020inflecting} report that encoder-decoders learn only the majority plural class in German (producing /-e/ for $\sim$80\% of pseudoword items vs. 45\% for humans; per-suffix $\rho = 0.05$--$0.33$, n.s.) and fail to extend suffixes productively, and \textcite{dankers2021generalising} show that an LSTM trained on German plurals encodes the same features (gender, final characters of the noun) a rule-based decision tree would but also uses a non-linguistic shortcut, using raw input length to predict the rare /-s/ class. Studies evaluating transformers on the past‑tense wug test similarly find weak correlation with human judgements \parencite{ma-gao-2022-get, corkery-etal-2019-yet}. These findings suggest that models capture phonological analogy effects but diverge from humans in ways that depend on the language, the inflection class, and the training regime. Unlike English suffixation or German plurals, the Spanish L-shaped morphome lacks phonological or semantic motivation, so models cannot rely on the cues those patterns provide. In what follows, we apply this model-human comparison to the Spanish L-shaped morphome: we replicate a human pseudoword study, evaluate transformers on the same items, and compare the two with the same analytical framework.

\section{Methods}
\label{sec:methods}

We first describe the human pseudoword study of \textcite{Nevins2015TheRA} that we replicate (Section \ref{subsec: human_conditions}) and the transformer models we evaluate against it (Section \ref{subsec: model_conditions}). The remaining subsections introduce three analyses applied identically to model predictions and human responses: match rate (Section \ref{subsec: match_rate_methods}), response preference (Section \ref{subsec: response_pref_methods}), and wordlikeness (Section \ref{subsec: wordlikeness_methods}).

\subsection{Human task}\label{subsec: human_conditions}

Our experiments build on the \textcite{Nevins2015TheRA}'s pseudoword study with Spanish speakers. Thirty pseudoword verbs were created: 15 \textit{L-shaped} (irregular) items, which instantiate the L-shaped pattern (the 1\textsc{sg.ind} stem recurring in all present subjunctive cells), and 15 \textit{NL-shaped} (regular) items, which do not. We use \textit{L} and \textit{NL} throughout to refer both to these two pseudoword types and to the two response types: an \textit{L-shaped response} uses the L-shape stem in the target cell, while an \textit{NL-shaped response} uses the regular alternative.   

\begin{figure}[ht]
    \centering
    \includegraphics[width=0.6\textwidth]{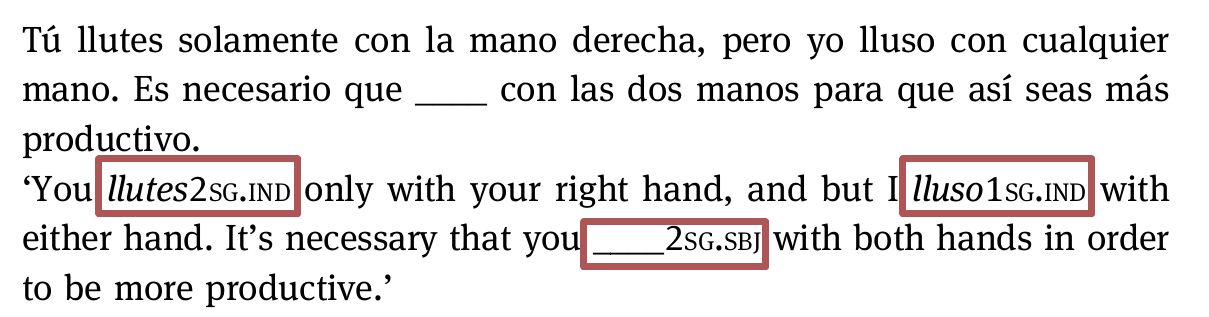}
            
            \caption{An example of pseudoword verb and its corresponding forms. \label{fig:stimuli-nevins-et-al}}
\end{figure}

The design is a six-cell paradigm in which only two cells are shown per trial. Participants are split into three groups:

\begin{itemize}
    \item \textbf{I$\rightarrow$S2:} given 1\textsc{sg.ind} and 2\textsc{sg.ind}, produce 2\textsc{sg.sbjv}.
    \item \textbf{I$\rightarrow$S3:} given 1\textsc{sg.ind} and 2\textsc{sg.ind}, produce 3\textsc{sg.sbjv}.
    \item \textbf{S$\rightarrow$I:} given 2\textsc{sg.ind} and 2\textsc{sg.sbjv}, produce 1\textsc{sg.ind}.
\end{itemize}

For pseudoword verbs where the two given cells share the same stem ((NL-shaped items)), the participant can produce the unseen form with that same stem. For pseudoword verbs where the two given cells differ in stem consonant (L-shaped items), the participant must choose which stem to extend to the unseen cell. The L-shaped pattern predicts that the 1\textsc{sg.ind} stem recurs in every present subjunctive cell, so a speaker who has internalized this morphome should produce the L-shaped stem in the unseen cell. The competing \emph{natural} response is grounded in morphosyntactically natural generalizations: Uniformity of Person in the I$\rightarrow$S groups (the 2\textsc{sg.ind} stem as the base for other 2\textsc{sg} forms) and Uniformity within Mood in the S$\rightarrow$I group (the 2\textsc{sg.ind} stem as the base for other indicatives). Random choice would indicate no systematic bias; systematic preference for the L-shaped stem on irregular items would indicate that speakers have internalized the morphome for novel items.

Because the L-shaped alternation rules are deterministic given any two cells of a pseudoword, \textcite{Nevins2015TheRA} reconstruct the full six-cell L-shaped paradigm for each item from the two given cells, yielding a single expected form for every remaining cell, including the target. This reconstructed paradigm provides the ground truth against which a response is scored as matching the L-shaped pattern or not: a matched response is one that reproduces the form the L-shaped paradigm predicts for that cell, and a non-matched response is one that does not.

The stimuli used invented stem alternations rather than real verbs, a design choice intended to minimize analogical interference from the Spanish lexicon. After preprocessing, \textcite{Nevins2015TheRA} retained 107 participants.

\subsection{Model task}\label{subsec: model_conditions}

To mirror the fill-in-the-blank design of Section \ref{subsec: human_conditions}, we frame the task as morphological re-inflection task: the model receives two filled paradigm cells and generates the target form. Spanish orthograhic forms were converted to IPA using an Argentinian Spanish orthography-to-phomeme mapping \parencite{ramarao-etal-2025-frequency}. For the example in Figure \ref{fig:stimuli-nevins-et-al}, the input is:

\texttt{\textipa{S} u t e s <V;IND;PRS;2;SG> \# \textipa{S} u s o <V;IND;PRS;1;SG> \# <V;SBJV;PRS;2;SG>}. 

We use the trained models from \textcite{ramarao-etal-2025-frequency}, which were trained from scratch on 39,435 Spanish verb samples under the three frequency conditions described in Section \ref{sec:intro}. We do not fine-tune or further train these models; we reuse the trained checkpoints as-is.

\subsubsection{Model architecture}

All models share a common encoder-decoder transformer backbone with 4 layers, 4 attention heads, embedding dimension $d=256$, and feedforward dimension 1024. Training used Adam \parencite{Kingma2015AdamAM} with a learning rate of 0.001, label smoothing of 0.1, and gradient clipping at 1.0, for 10,000 update steps with checkpoints saved every 10 epochs. Inference uses beam search with width 5. For each of the three frequency conditions, 12 independent checkpoints were trained.

\subsection{Match rate analysis}\label{subsec: match_rate_methods}

We score each model prediction against the L-shape-reconstructed target described in Section \ref{subsec: human_conditions}. We report two metrics:

\textbf{Overall match rate.} The proportion of predictions that exactly match the
reconstructed target form.

\textbf{Stem match rate.} The proportion of predictions in which the expected L-shaped stem appears, regardless of the rest of the form. 

\subsection{Response preference analysis}\label{subsec: response_pref_methods}

Match rate alone does not capture the full picture of model behavior \parencite{Elsner2019}: it tells us how often a prediction matches the target, but not which alternative form the model produced when it did not. To examine the distribution of response preferences for L-shaped vs. NL-shaped (regular) responses across participants, models, and items, we apply the same analytical framework as \textcite{Nevins2015TheRA}: log(NL/L) response ratios computed by group (across participants or model checkpoints) and by item (across pseudoword stimuli).

For each participant or model checkpoint, we compute the log-ratio $\log(\text{NL}/\text{L})$ of NL-shaped to L-shaped responses and visualize the density across participants and across model checkpoints. Positive values indicate a preference for NL-shaped (regular) responses; negative values indicate an L-shaped preference; zero is neutral.

We use two complementary tests on the per-item log-ratios. Spearman rank correlation asks whether two groups share the same ordering of items, independently of absolute magnitudes. The two-sample Kolmogorov-Smirnov test instead compares the full distributional shape, with a larger $D$ indicating a larger distributional gap.

We run both tests in two configurations: model-vs-human and model-vs-model.

\subsection{Wordlikeness analysis}\label{subsec: wordlikeness_methods}

Phonological similarity to existing words is a well-established driver of morphological generalization in humans \parencite{albright2007natural, TangBH2023, olejarczuk2018distributional}. Item-level variation could still be explained by such lexical analogy, so we compare the pseudoword items of \textcite{Nevins2015TheRA} to real Spanish L-shaped verb stems and test whether the similarity predicts response patterns in participants and models.

We operationalize L-shaped wordlikeness following \textcite{TangBH2023}, using the Generalized Neighborhood Model \parencite{Nosofsky1986, bailey2001determinants}. GNM quantifies cross-lexicon similarity through phonological neighborhood density and controls for lexical frequency by assigning equal frequency weights to all items. Psychological distances between test items and the reference lexicon (all Spanish L-shaped verb stems) use a cost matrix derived from the PHOIBLE feature table \parencite{StevenMoran2019}; substitution cost is the proportion of mismatched features, and a distance decay \parencite{albright2007natural} down-weights more distant neighbors. 

The reference lexicon contains the L-shaped stems of real Spanish L-shape verbs (e.g., \textit{dig-} from the alternation \textit{decir}/\textit{digo}); the test lexicon contains the L-shaped stems of the artificial pseudoword paradigms (e.g., \textit{but-}, \textit{prafit-}). For each test stem, GNM returns a single wordlikeness score: the sum, over all reference stems, of distance-decayed similarities, so that closer and more numerous neighbors yield a higher score. We then pair each GNM score with the trial-level response data (whether the response on that trial was L-shaped or NL-shaped) and partition the responses into four subsets: human participants and each of the three model conditions (10\%L-90\%NL, 50\%L-50\%NL, 90\%L-10\%NL). 

We expect higher L-shaped wordlikeness to predict more L-shaped responses: a negative slope between GNM and the per-item log(NL/L) ratio (more L-shape lowers the ratio), and a positive $\beta$ on wordlikeness in a trial-level model of an L-shaped answer.

We run two complementary analyses to test this. The first is a per-item scatter plot: each pseudoword contributes one point per group, with GNM wordlikeness on the x-axis and the per-item log-ratio (from Section \ref{subsubsec: density_response_preference_by_item}) on the y-axis, plus a linear fit per group. The second is a mixed-effects logistic regression that tests the same relationship at the trial level, asking whether wordlikeness predicts an L-shaped answer with proper inferential uncertainty (a coefficient and $p$-value), while random intercepts absorb item-level and participant-level variation. The two analyses can disagree because they sit at different levels of aggregation; reporting both lets us see the item-level pattern and whether it survives a trial-level test.

We fit this regression with the \texttt{glmer} function from the \texttt{lme4} package \parencite{lme4} in \texttt{R}. The dependent variable is \texttt{answer\_choice} (L as the positive class vs. NL), the fixed effect is L-shaped wordlikeness, and the random intercepts are over \texttt{item} and participant:

\begin{center}
\texttt{
answer\_choice $\sim$ L-shaped\_word\_likeness + (1|item) + (1|participant)
}
\end{center}

We fit four regressions in total: one for human participants and one for each of the three model conditions (10\%L-90\%NL, 50\%L-50\%NL, 90\%L-10\%NL). For the human regression, \texttt{(1|participant)} ranges over 107 participants; for each regression on a model condition, the same random intercept ranges over the 12 checkpoints in that frequency condition. The human regression is fit on 749 trial-level observations, while each model-condition regression is fit on approximately 8,000 observations (12 checkpoints $\times$ trials on L-shaped pseudoword items).

\section{Results}
\label{sec:results}

We first evaluate how well models match human performance on match rate metrics, then examine the distribution of response preferences across participants, models, and items, and finally test whether phonological similarity to real Spanish L-shaped verbs explains item-level variation. 

\subsection{Match rate}\label{subsec: match_rate_results}

\subsubsection{Overall match rate}\label{subsubsec: overall_accuracies}

In this section, we assess overall match rate against the reconstructed L-shaped target form on the \textcite{Nevins2015TheRA} test items. Human participants reached an overall match rate of 21.64\% (SD=22.48\%, 95\% CI [17.38, 25.90]). All three model conditions sit within this range: 10\%L-90\%NL at 20.82\% (SD=7.75\%, CI [16.44, 25.20]), 50\%L-50\%NL at 18.54\% (SD=8.03\%, CI [14.00, 23.09]), and 90\%L-10\%NL at 25.29\% (SD=5.35\%, CI [22.27, 28.32]). Full-form reconstruction is therefore difficult for both humans and models, and all three model conditions fall within the human range. The 90\%L-10\%NL condition is the only one whose own confidence interval sits entirely above the human mean, consistent with its training bias toward L-shaped forms that coincide with the reconstructed (L-shape) targets.

\subsubsection{Stem match rate}\label{subsubsec: stem_accuracies}

Human participants reached a stem match rate of 32.60\% (SD=22.39\%, 95\% CI [28.36, 36.85]). All three model conditions exceed this: 10\%L-90\%NL at 49.43\% (SD=2.99\%, CI [47.73, 51.12]), 50\%L-50\%NL at 43.61\% (SD=4.57\%, CI [41.03, 46.19]), and 90\%L-10\%NL at 43.07\% (SD=4.55\%, CI [40.49, 45.64]). The small spread across training conditions (43--49\%) indicates that stem learning is robust to variations in the proportion of L-shaped verbs in training. Stem match rate exceeds overall match rate for all three model conditions, showing that models select the right stem even when they fail at full-form reconstruction.

\subsection{Response preference}\label{subsec: response_preferences}

\subsubsection{By group}

Figure \ref{plot: response_preference_by_participant} shows the density of log(NL/L) response ratios. Participants' mean ratio is 0.62. Models show the opposite tendency: 10\%L-90\%NL at $-$0.08 and 50\%L-50\%NL at $-$0.09 hover near neutrality, while 90\%L-10\%NL at $-$0.29 shows a clear L-shaped preference. Humans and models diverge most in the 90\%L condition. The densities also differ in shape: the model curves are narrow and peaked, while the human curve is wide and flat. Per-participant ratios are inherently noisier than per-checkpoint ones (see Limitations).

\begin{figure}[ht]
  \centering      
    \includegraphics[width=0.6\linewidth]{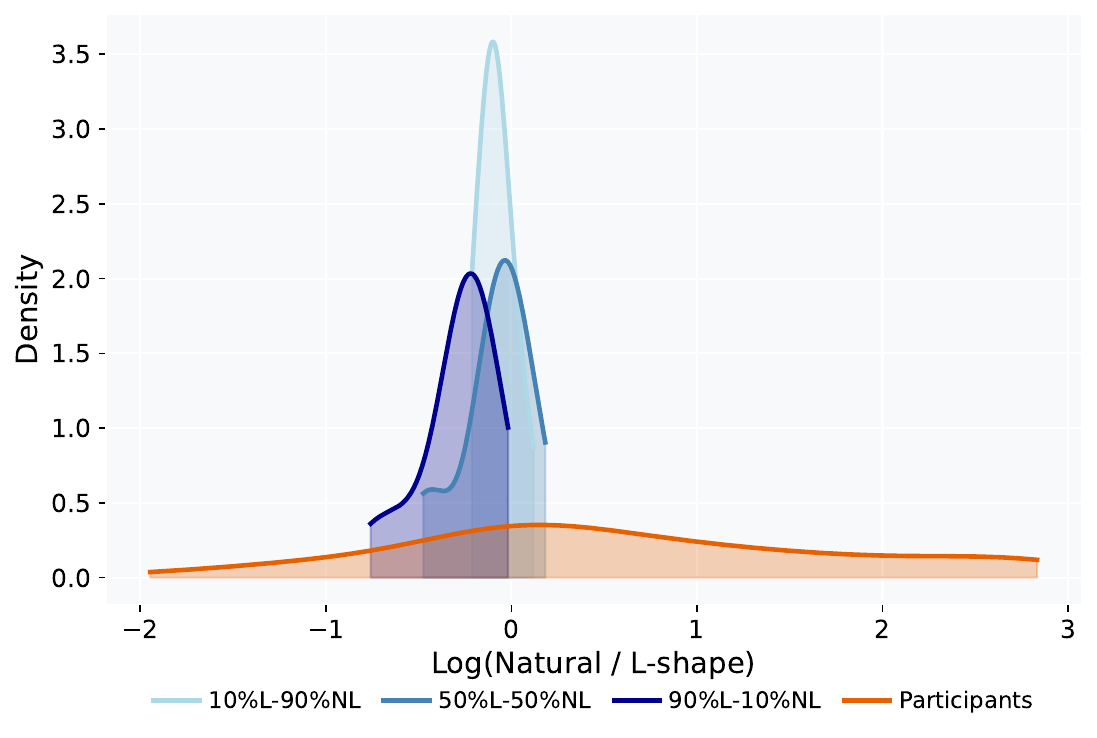}
    \caption{Density of $\log(\text{NL}/\text{L})$ response ratios per participant and per model checkpoint. Positive values indicate an NL-shaped (regular) preference; negative values indicate an L-shaped preference. The dashed vertical line marks neutral preference.\label{plot:
  response_preference_by_participant}}
\end{figure}

\subsubsection{By item}\label{subsubsec: density_response_preference_by_item}

\begin{figure}[ht]
\centering                                                                                 
\includegraphics[width=0.6\linewidth]{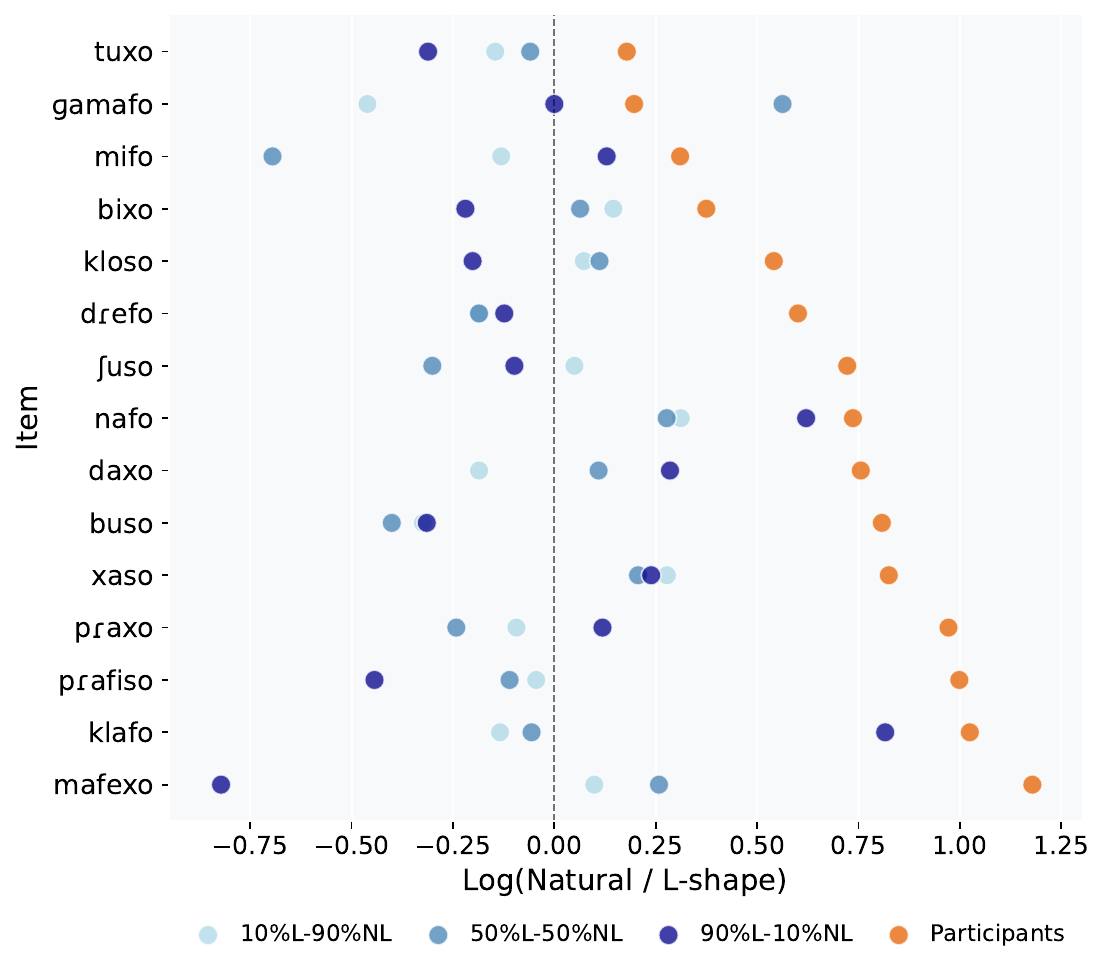}                     
\caption{Per-item response preference ($\log(\text{NL}/\text{L})$) for participants and the three model conditions. Positive values indicate an NL-shaped (regular) preference; negative values indicate an L-shaped preference. The dashed vertical line marks neutral preference.\label{plot:
response_preference_models_participants}}
\end{figure} 

Figure \ref{plot: response_preference_models_participants} shows the per-item response preferences. Human participants consistently preferred NL-shaped forms across all items, with a mean log-ratio of 0.68. The strongest NL-shaped preferences are for [\textipa{mafexo}] (1.18), [\textipa{klafo}] (1.02), and [\textipa{pRafiso}] (1.00); the weakest are [\textipa{tuxo}] (0.18), [\textipa{gamafo}] (0.20), and [\textipa{mifo}] (0.31).

In the 10\%L-90\%NL condition, the mean log-ratio is $-$0.05, a slight overall preference for L-shape. Items with stronger NL-shaped preferences include [\textipa{nafo}] (0.31), [\textipa{xaso}] (0.28), and [\textipa{bixo}] (0.15); items with stronger L-shaped preferences include [\textipa{gamafo}] ($-$0.46), [\textipa{buso}] ($-$0.32), and [\textipa{dRefo}] ($-$0.19).

In the 50\%L-50\%NL condition, the mean is $-$0.03, near-neutral. Items with stronger NL-shaped preferences include [\textipa{gamafo}] (0.56), [\textipa{nafo}] (0.28), and [\textipa{mafexo}] (0.26); items with stronger L-shaped preferences include [\textipa{mifo}] ($-$0.69), [\textipa{buso}] ($-$0.40), and [\textipa{Suso}] ($-$0.30).

In the 90\%L-10\%NL condition, the mean is $-$0.02, also near-neutral. Items with stronger NL-shaped preferences include [\textipa{klafo}] (0.82), [\textipa{nafo}] (0.62), and [\textipa{daxo}] (0.29); items with stronger L-shaped preferences include [\textipa{mafexo}] ($-$0.82), [\textipa{pRafiso}] ($-$0.44), [\textipa{buso}] ($-$0.31), and [\textipa{tuxo}] ($-$0.31).

Humans consistently prefer NL-shaped responses across all items; models show a tendency toward L-shaped responses that correlates with the frequency of L-shaped verbs in training. Some items, such as [\textipa{nafo}] and [\textipa{xaso}], elicit NL-shaped preferences across all frequency conditions.

\paragraph{Correlation and distributional similarity.}\label{subsubsec: correlation}

Although humans and models diverge in mean preference, they could still agree on which items elicit relatively more NL- versus more L-shaped responses, or share the same distributional shape with different means. The following two analyses test these possibilities.

None of the models show statistically significant correlation with human rankings ($p$s $>$ 0.05), but the 10\%L-90\%NL condition, which most closely approximates the natural distribution, has the highest correlation ($\rho$ = 0.25) (Table \ref{tab: spearman-rank-correlation-models-participants}).

All three models differ significantly from participants in distributional shape ($p$s $<$ 0.001; Table~\ref{tab: ks-model-participants}). The 90\%L-10\%NL model shows the largest gap ($D = 0.72$), followed by 10\%L-90\%NL ($D = 0.63$) and 50\%L-50\%NL ($D = 0.58$). Among models, 10\%L-90\%NL and 50\%L-50\%NL are closest to each other ($D = 0.25$, $p = 0.86$); both differ significantly from 90\%L-10\%NL ($D = 0.67$ and $D = 0.58$ respectively).

\begin{table}[ht]
    \centering
    \small
    \setlength{\tabcolsep}{3pt}
    \begin{minipage}[t]{0.4\columnwidth}
        \centering             

        \begin{tabular}{lrr}
        \hline
        Comparison & $\rho$ & \textit{p}-val. \\               
        \hline             
          10\%L-90\%NL vs. Participants & 0.25 & 0.37 \\
          50\%L-50\%NL vs. Participants & 0.03 & 0.91 \\                
          90\%L-10\%NL vs. Participants & 0.02 & 0.95 \\         
          \hdashline
          10\%L-90\%NL vs. 50\%L-50\%NL & 0.33 & 0.23 \\        
          10\%L-90\%NL vs. 90\%L-10\%NL & 0.05 & 0.86 \\         
          50\%L-50\%NL vs. 90\%L-10\%NL & 0.15 & 0.62 \\         
          \hline        
  \end{tabular}                          

  \caption{Spearman rank correlation between models and participants (rows 1--3) and between   
  models (rows 4--6).\label{tab: spearman-rank-correlation-models-participants}}               
  \end{minipage}
  \hfill                                      
  \begin{minipage}[t]{0.5\columnwidth}
    \centering
    \small
    \setlength{\tabcolsep}{3pt}
    \begin{tabular}{lrr}                        
        \hline
        Comparison & D & \textit{p}-val. \\
        \hline           
          10\%L-90\%NL vs. Participants & 0.63 & $<$ 0.001$^{***}$ \\
          50\%L-50\%NL vs. Participants & 0.58 & $<$ 0.001$^{***}$ \\              
          90\%L-10\%NL vs. Participants & 0.72 & $<$ 0.001$^{***}$ \\
          \hdashline
          10\%L-90\%NL vs. 50\%L-50\%NL & 0.25 & 0.86 \\        
          10\%L-90\%NL vs. 90\%L-10\%NL & 0.67 & $<$ 0.01$^{**}$ \\         
          50\%L-50\%NL vs. 90\%L-10\%NL & 0.58 & $<$ 0.05$^{*}$ \\
        \hline             
    \end{tabular}

  \caption{Kolmogorov-Smirnov D-statistics between models and participants (rows 1--3) and     
  between models (rows 4--6). $^{*}\,p \leq 0.05$, $^{**}\,p \leq 0.01$, $^{***}\,p \leq
  0.001$.\label{tab: ks-model-participants}}
  \end{minipage}  
\end{table}

On both measures, no model matches the human pattern: the highest rank correlation reaches only $\rho = 0.25$, and all three distributions differ significantly from the human one. Among the three, the 10\%L-90\%NL model shows the closest alignment with human preferences on both metrics.

\subsection{Wordlikeness}\label{subsec: wordlikeness_results}

\begin{figure}[h]
  \centering
    \includegraphics[width=0.6\columnwidth]{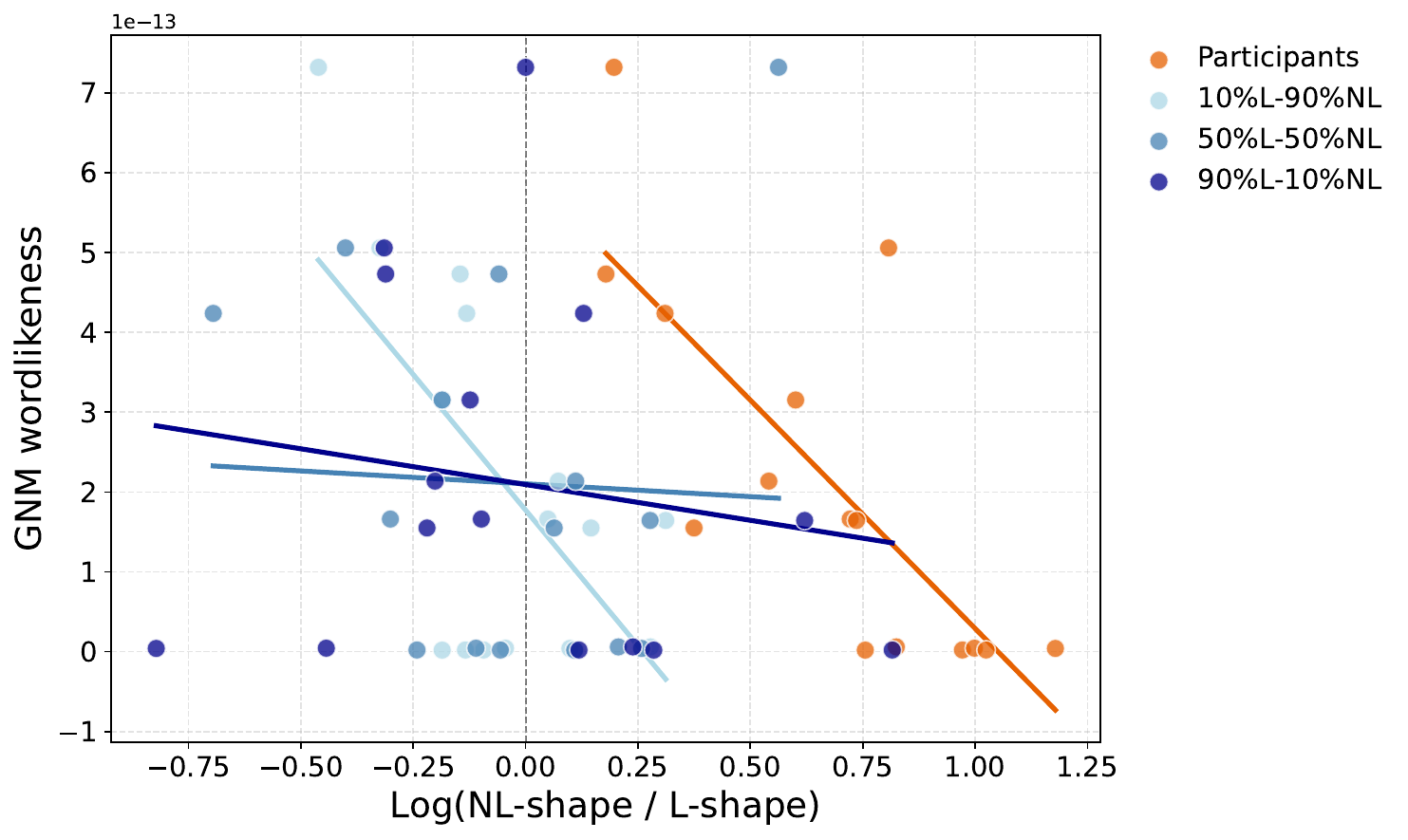}
    \caption{L-shaped wordlikeness (GNM) vs. per-item response preference $\log(\text{NL}/\text{L})$ for participants and the three model conditions. Positive values on the x-axis indicate an NL-shaped (regular) preference; negative values indicate an L-shaped preference. Coloured lines are linear fits per group; the dashed vertical line marks neutral preference.\label{plot: scatter-combined}}
\end{figure}

Figure \ref{plot: scatter-combined} shows the per-item GNM wordlikeness against response preference for participants and the three model conditions. All four show a negative correlation; humans and 10\%L-90\%NL are more correlated than 50\%L-50\%NL and 90\%L-10\%NL.

For humans, L-shaped wordlikeness does not have a significant effect ($\beta = -0.147$, $p =0.118$). For models, we find a significant positive effect in the 10\%L-90\%NL condition ($\beta = 15.085$, $p < 0.001$) and in the 50\%L-50\%NL condition ($\beta = 2.784$, $p < 0.05$). The 90\%L-10\%NL condition shows no effect ($\beta = -1.07$, $p = 0.226$).

The more realistic model, 10\%L-90\%NL is sensitive to phonological analogy in the expected direction; the least realistic one (90\%L-10\%NL) is not. This lexical analogy effect is consistent with prior psycholinguistic findings \parencite[see][and references therein]{TangBH2023, olejarczuk2018distributional}. Humans, surprisingly, are not sensitive to it in the \textcite{Nevins2015TheRA} study, despite the  correlation in Figure \ref{plot: scatter-combined} (see Limitations).

\section{Discussion}
\label{sec:discussion}

Both humans and models reached only moderate stem match rate on pseudoword items (humans 33\%; models 43--49\%). This is in line with prior findings that neural models perform poorly on irregular inflection in English past tense \parencite{ma-gao-2022-get} and German plurals \parencite{liu-hulden-2022-transformer}, with transformers' failure to productively generalize this same L-shaped morphome to novel forms \parencite{ramarao-etal-2026-character}, and more broadly with the evidence that  transformer accuracy drops sharply when generalizing to novel lemmas \parencite{Goldman2022UnsolvingMI, kodner2023morphological}.

A clearer divergence emerges in response preferences. Humans consistently favored NL-shaped forms across all items, treating the L-shaped as a rare exception. Models showed the opposite tendency: their preference for L-shaped forms grew with the proportion of L-shaped verbs in training. Alignment with human behavior was weak overall, but the model trained on the naturalistic distribution (10\%L-90\%NL) had the highest rank correlation with Nevins's Spanish speakers. The L-shaped morphome is thus more productive in our models than in Nevins's participants (\~33\% L-shape extension), and the model pattern actually sits closer to that reported by \textcite{cappellaro2024cognitive} for Italian, where 60\% of participants selected the L-shaped alternative. Two factors may explain both the Nevins/Cappellaro gap and why our models pattern with Cappellaro: (i) the L-shape may be more productive in Italian than in Spanish (Cappellaro's own argument), and (ii) Cappellaro's task is forced choice between two given forms while Nevins's is open production, which is much harder. Beam-search generation puts our models closer to forced choice than to production, since it always returns the single most probable form rather than the full distribution of plausible responses. 

Models trained on the naturalistic and balanced distributions were sensitive to phonological similarity between pseudoword items and real Spanish L-shaped verb stems, they were more likely to produce an L-shaped form when the test item resembled a real L-shaped verb. Human participants were not sensitive to this factor. The 90\%L-10\%NL condition, which is the least  natural distribution, was furthest from human participants on both preference distributions and sensitivity to lexical analogy. This finding speaks to the \textcite{Nevins2015TheRA}--\textcite{cappellaro2024cognitive} debate. \textcite{cappellaro2024cognitive} argue that \textcite{Nevins2015TheRA}'s null result reflects the design choice, in particular, the use of phonologically unfamiliar stem alternations that block analogy to real L-shaped verbs, rather than speakers having failed to internalize the morphome. Our models trained on naturalistic and balanced distributions shows the analogical mechanism Cappellaro claim is central to morphomic activation: they produce more L-shaped responses for items that resemble Spanish L-shaped verbs.

\section{Conclusion}
\label{sec:conclusion}

In sum, the L-shaped morphome is learnable from distributional input by encoder-decoder transformers trained from scratch on Spanish verb inflection, but the resulting generalization diverges qualitatively from human behavior. The frequency manipulation proved informative: models trained on more realistic distributions behaved more like humans in their sensitivity to phonological analogy, even as their overall response preferences remained fundamentally different. 

\section*{Limitations}\label{sec:limitations}

The human participants were tested on stimuli with sentence context (Figure \ref{fig:stimuli-nevins-et-al}), while the models were not (Section \ref{subsec: model_conditions}). Any differences between models and humans could therefore arise from this experimental difference. 

The higher variation in responses by the human participants compared to the models (Figure \ref{plot: response_preference_by_participant}) might be due to the fact that the number of human participants is 10 times higher than the number of models (107 vs 12). Furthermore, the sample size of the human responses is 10 times smaller than that of the models ($\approx$ 750 vs 8,100). Together, they might explain the null effect of wordlikeness found with the human responses in the regression model.

We investigated only the wordlikeness of the pseudowords in terms of how they resemble the L-shaped verbs and not the natural verbs. Furthermore, we did not use a different lexicon that each model trained on to estimate a separate set of wordlikeness for each model, but rather we estimated over all L-shaped verbs.

\section*{Data availability}\label{sec:data-availability}

The code and data supporting this study, including the model predictions, participant responses, and all analysis scripts, are available in an anonymous public repository: \url{https://anonymous.4open.science/r/modeling-nevins-et-al-43D0}

\printbibliography

\end{document}